\documentclass[preprint,authoryear,1pt]{elsarticle}

\makeatletter
\newif\if@restonecol
\makeatother

\usepackage{graphicx,amssymb}
\usepackage[linesnumbered,ruled,vlined]{algorithm2e}
\usepackage{amsmath}
\usepackage[switch,pagewise]{lineno}
\usepackage[usenames]{color}
\usepackage{longtable}
\usepackage{booktabs}
\usepackage{url}
\usepackage{mathrsfs}
\usepackage{amsfonts}
\usepackage{marvosym}
\usepackage{lscape}
\usepackage{enumerate}
\usepackage{caption}
\usepackage{subcaption}
\usepackage{multirow}
\usepackage{verbatim}
\usepackage{ulem}

\begin{document}

\begin{frontmatter}

\title{Combining tabu search and graph reduction to solve the maximum balanced biclique problem}

\author[label1]{Yi Zhou} \ead{zhou@info.univ-angers.fr} 
\author[label1,label2]{Jin-Kao Hao\corref{cor}} \ead{jin-kao.hao@univ-angers.fr} \cortext[cor]{Corresponding author}
\address[label1]{LERIA, University of Angers, 2 bd Lavoisier, 49045 Angers, France}
\address[label2]{Institut Universitaire de France, 1 rue Descartes, 75231 Paris, France}

\begin{abstract}
The Maximum Balanced Biclique Problem is a well-known graph model with relevant applications in diverse domains. This paper introduces a novel algorithm, which combines an effective constraint-based tabu search procedure and two dedicated graph reduction techniques. We verify the effectiveness of the algorithm on 30 classical random benchmark graphs and 25 very large real-life sparse graphs from the popular Koblenz Network Collection (KONECT). The results show that the algorithm improves the best-known results (new lower bounds) for 10 classical benchmarks and obtains the optimal solutions for 14 KONECT instances.

\end{abstract}
\end{frontmatter}

\section{Introduction}
\label{sec_introduction}

Given a bipartite graph $G=(U,V,E)$ with two disjoint vertex sets $U$, $V$ and an edge set $E\subseteq U\times V$, a biclique $(X,Y) = X\cup Y$ is the union of two subsets of vertices $X\subseteq U$, $Y\subseteq V$ such that $u\in X, v\in Y$ implies that $\{u,v\}\in E$. In other words, the subgraph induced by the set of vertices $X\cup Y$ is a complete bipartite graph. If $|X|=|Y|$, then $(X,Y)$ is called a balanced biclique. The Maximum Balanced Biclique Problem (MBBP) is to find a balanced biclique $(X^*,Y^*)$ of maximum cardinality of $G$, $(X^*,Y^*)$ being the maximum balanced biclique of size $|X^*|$ (or $|Y^*|$) \citep{garey1979computers}.

As shown in \citep{dawande2001bipartite}, by following a well-known integer linear programming model of the more general maximum vertex weight biclique problem, MBBP can be formulated as a binary linear program as follows.

\begin{equation}\label{f1}
\max \ \ \omega(G)=\sum_{i=1}^{|U|}{x_i}
\end{equation}
subject to:
\begin{equation}\label{c1}
x_i + x_j \leq  1, \forall\{i,j\}\in \bar{E}
\end{equation}
\begin{equation}\label{c2}
\sum_{i=1}^{|U|}{x_i} - \sum_{j=|U|+1}^{|U|+|V|}{x_j} = 0
\end{equation}
\begin{equation}\label{c3}
x_i, x_j \in \{0, 1\}
\end{equation}

\noindent where each vertex of $U\cup V$ is associated to a binary variable $x_i$ indicating whether the vertex is part of the biclique, $\bar{E}$ is the set of edges in the bipartite complement of $G$. Objective (\ref{f1}) maximizes the size of the biclique. Constraint (\ref{c1}) ensures that each pair of non-adjacent vertices cannot be selected at the same time (i.e., the solution must be a clique). Equation (\ref{c2}) enforces that the returned biclique is balanced.

In terms of computational complexity, the decision version of MBBP is NP-complete in the general case \citep{garey1979computers,alon1994algorithmic}, even though the maximum biclique problem without the balance constraint (Eq. (\ref{c2})) is polynomially solvable by the maximum matching algorithm \citep{cheng2000biclustering}.

MBBP is a prominent model with many applications. For example, in nanoelectronic system design, MBBP is used to identify the maximum defect-free crossbar from a partially fabricated defective crossbar represented by a bipartite graph \citep{tahoori2006application,al2007defect}. In computational biology, MBBP is applied to simultaneously group genes and their expressions under different conditions (called biclustering) \citep{cheng2000biclustering}. Another application can be found in the field of VLSI for PLA-folding \citep{ravi1988complexity}. Generally, the clique and biclique models are also popular tools for winner determination in combinatorial auctions \citep{WuHao2015wdp}, tail dependence structure analysis of the foreign exchange market \citep{WangXie2016exchangemarket},  co-location feature pattern mining in space \citep{Yanetal2015colocation}.


Given the significance of MBBP as a NP-hard problem and its relevance in practice, a number of methods, including approximate, exact and heuristic algorithms have been proposed and investigated in the literature. For example, in \citep{feige2004hardness}, the relations between the approximate hardness of MBBP and 3-SAT as well as the maximum clique problem were established. In \citep{mubayi2010finding}, despite the NP-hardness of MBBP, a polynomial algorithm was given to find a balanced biclique with size $\lfloor{\frac{\ln{n}}{\ln{(2en^2/m)}}}\rfloor $ (the cardinality of $|X|$ or $|Y|$) for a graph with $n$ vertices and $m$ edges. In \citep{tahoori2006application}, a recursive exact algorithm for searching a maximum balanced biclique with a given size was proposed. However, the computational time of this algorithm becomes prohibitive when the number of vertices of the graph exceeds $(32,32)$. In \citep{mccreesh2014exact}, another exact approach for MBBP for general (non-bipartite) graphs was studied. This algorithm follows the classical branch and bound framework for the popular maximum clique problem \citep{wu2015review} with additional symmetry breaking techniques.

To cope with the computational challenge of MBBP, heuristic methods constitute an interesting approach. These methods aim to obtain satisfactory solutions in an acceptable time frame without guaranteeing the optimality of the attained solutions. From an algorithmic point of view, rather than directly seeking the maximum balanced biclique in the given graph, the majority of existing heuristic algorithms solved the equivalent maximum balanced independent set problem for the bipartite complement. For example, \cite{tahoori2006application} proposed a greedy heuristic algorithm based on vertex-deletion, which iteratively removes vertices with maximum degree from the bipartite complement until the set of remaining vertices forms an independent set (i.e., a set of vertices such that no edge exists between any pair of vertices in the graph). \cite{al2007defect} presented an improved greedy heuristic, in which the vertex connecting the maximum number of vertices of minimum degree is removed. \cite{yuan2011low} introduced another greedy heuristic algorithm, which iteratively deletes vertices adjacent to the maximum number of vertices in a restricted set. Then \cite{yuan2014fast} accelerated this algorithm by removing multiple vertices at each iteration. Recently, \cite{yuan2015new} proposed a powerful (and rather complex) evolutionary algorithm combining structure mutation and repair-assisted restart. The computational results showed that this algorithm performed very well on random dense graphs, which represent one type of the most challenging instances for MBBP. We will use this algorithm as one of the main references for our comparative studies.

On the other hand, graphs from real-life applications like social networks and biological networks are usually very large with millions even billions of vertices, rendering most existing approaches unpractical. In this study, we aim to fill the gap by developing improved methods for MBBP, which should be able to handle both random dense graphs and very large real-life networks. Based on an analysis of the studied problem and existing algorithms (Section \ref{Rationale}), we introduce a new and effective algorithm named tabu search with graph reduction for MBBP (TSGR-MBBP), which combines an original Constraint-Based Tabu Search (CBTS) and two dedicated graph reduction techniques. We identify the main contributions of this study as follows. 

\begin{enumerate}
	\item From an algorithmic perspective, the proposed TSGR-MBBP algorithm seeks maximum balanced bicliques directly on the given graph. Compared to the existing approaches which search for balanced independent sets on the complement, operating on the given graph has an advantage of requiring less memory for large sparse graphs. More importantly, TSGR-MBBP employs the Constraint-Balanced Tabu Search algorithm to effectively explore the search space (Section \ref{sec_tabu_search}) and two bound-based dedicated reduction techniques to shrink progressively the given graph (Sections \ref{subsec_peel} and \ref{subsec_exact_search}). This is the first study combining local optimization and graph reduction within the iterated search framework for MBBP.

	\item We demonstrate the effectiveness of the proposed algorithm on two sets of 55 MBBP benchmark instances (Section \ref{sec_experiments}). For the set of 30 random challenging instances, the algorithm dominates state-of-the-art algorithms including the current best-performing algorithm presented in \citep{yuan2015new} and the powerful mixed integer programming solver CPLEX. The algorithm also obtains 10 improved best solutions (i.e., new lower bounds) and matches the best-known results for the remaining 20 instances. For the 25 very large real-life instances from the well-known Koblenz Network Collection, the algorithm proves, for the first time, the optimal solutions for 14 instances (by obtaining the same upper and lower bounds) and obtains tight lower bounds (better than those of CPLEX) for the remaining instances. We also show an analysis of key components (CBTS and the reduction methods) to get insight of their usefulness (Section \ref{sec_analysis}).

\end{enumerate}

The remainder of the paper is organized as follows. Section \ref{sec_prob_formul} provides some useful notations. Section \ref{sec_irs_algo} introduces the proposed algorithm. Computational results on benchmark instances are presented in Section \ref{sec_experiments}. Section \ref{sec_analysis} shows an analysis of the key components of the proposed algorithm, followed by conclusions in the final section.

\section{Preliminary definitions}
\label{sec_prob_formul}

Let $G=(U,V,E)$ be a bipartite graph, we introduce the following notations and definitions which are needed for the description of the proposed approach.

\begin{itemize}
\item [-] Given a vertex $v \in  U\cup V$, $N(v)$ denotes the set of vertices adjacent to $v$, i.e., $N(v)=\{u : \{v,u\}\in E\}$. Clearly, if $v \in  U$, then $N(v) \subseteq  V$, otherwise, $N(v) \subseteq U$.

\item [-] Given $S\subseteq U\cup V$, $N(S)$ denotes the subset of vertices from $(U\cup V)\setminus S $ that are adjacent to at least one vertex in $S$, i.e., $N(S)= (\bigcup\limits_{i \in S} N(i)) \setminus S$. 

\item [-] Given $X\subseteq U$, $Y \subseteq V$, $G[X\cup Y]=(X,Y,E(X\cup Y))$ denotes the subgraph induced by $X\cup Y$. If $G[X\cup Y]$ is a complete bipartite graph, i.e., $E(X,Y) = X \times Y$, then $X\cup Y$ is a biclique, which is also denoted by $(X,Y)$.

\item[-] Given a biclique $(X,Y)$, the balanced size of $(X,Y)$ is $min(|X|,|Y|)$, and the balance deviation is $||X|-|Y||$. If the balance deviation is 0, $(X,Y)$ is a balanced biclique of size $|X|$ (or $|Y|$).

\end{itemize}

	\begin{figure}
	\centering\scalebox{0.8}{\includegraphics[scale=0.55]{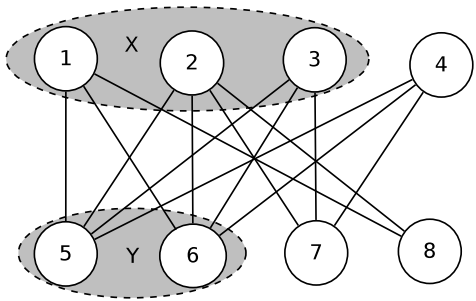}}
	\centering\caption{$X=\{1,2,3\}$, $Y=\{5,6\}$, $N(X\cup Y)=\{4,7,8\}$, $(X,Y)$ is a biclique of balanced size 2. The balance deviation of $(X,Y)$ is 1.} \label{fig_demo1}
	\end{figure}		
	
Figure \ref{fig_demo1} illustrates the above definitions with a bipartie graph composed of 8 vertices and 13 edges.

Let $\Omega(G)$ denote the search space composed of all balanced bicliques in $G$, $\Omega^k$ be the relaxed search space including all bicliques with a balance deviation no more than $k$ ($k\geq 0$), i.e., $\Omega^k = \{(X,Y):X \subseteq U, Y\subseteq V, E(X,Y) = X \times Y, ||X|-|Y||\leq k\}$, then, as explained in the next section, our algorithm explores bicliques in the (slightly) relaxed search space $\Omega^2$ (i.e., with a balance deviation limited to 2) rather than the search space of strictly balanced bicliques $\Omega(G)$ (i.e., $\Omega^0$). 

Finally, the quality of a biclique $(X,Y)$ in $\Omega^k$ is measured by its balanced size $min(|X|,|Y|)$. Given two bicliques $(X_1,Y_1)$ and $(X_2, Y_2)$, $(X_1,Y_1)$ is better than $(X_2,Y_2)$ if $min(|X_1|,|Y_1|) > min(|X_2|,|Y_2|)$.

\section{Tabu search with graph reduction}
\label{sec_irs_algo}

This section introduces our proposed TSGR-MBBP algorithm for solving the maximum balanced biclique problem. TSGR-MBBP is based on the well-known tabu search metaheuristic \citep{glover1997tabu}, which is specifically adapted to the MBBP problem. Indeed, tabu search being a general method, it is critical to find a suitable adaptation of the method to the considered problem (e.g., \citep{Daiazetal2017,Elhedhlietal2014,MaWangHao2017}). For our MBBP problem, we proposed the constraint-based tabu search (Section \ref{sec_tabu_search}), which is able to effectively explore the search space of slightly unbalanced bi-cliques. In order to help tabu search to avoid non-promising search regions, we introduce two bound-based graph reduction techniques to shrink progressively the given graph (Sections \ref{subsec_peel} and \ref{subsec_exact_search}).

\subsection{Rationale of the proposed approach}
\label{Rationale}

Many real-life networks have millions or even billions of vertices with a very low edge density. Existing approaches for solving MBBP rely heavily on the complement and the adjacent matrix representation. Unfortunately, the complement of such a massive graph usually results in very high memory consumption, making most of existing MBBP approaches unpractical. To avoid this difficulty, the proposed algorithm operates directly on the given graph, implying that much less memory is required for processing very large real-life sparse networks. From an algorithmic perspective, our algorithm iteratively seeks improved solutions by local search combined with graph reduction strategies. Specifically, the algorithm starts from an initial solution (a slightly relaxed balanced biclique) and uses move operators to improve the solution iteratively. However, we still need to answer a crucial question: how to improve the solution effectively while maintaining the two main constraints of a solution (balanced and biclique)? 

Intuitively, local search operators that are successful for the maximum clique problem \citep{wu2015review} can be applied to MBBP, such as ``add" (adding a vertex to the solution), ``swap" (exchanging a vertex in the solution with another vertex out of the solution) or still ``drop" (dropping a vertex from the solution). However, given a balanced biclique, an application of any of these operators results in an unbalanced biclique. To cope with this difficulty, we propose to (slightly) relax the balance requirement of the solution and allow the algorithm to explore both balanced and slightly unbalanced bicliques. For this purpose, we adopt the generalized ``push" operator initially designed for the maximum vertex weight clique problem \citep{Zhouetal2017} to explore solutions within the relaxed search space $\Omega^2$ rather than $\Omega^0$.

Another key idea we used is graph reduction. Given a bipartite graph $G$ and a known best balanced size $\omega$ (a lower bound), it is clear that to further improve $\omega$, it is useless to consider any vertex whose degree is smaller than or equal to $\omega$ since such a vertex can in no way extend the best solution found so far. Consequently, these vertices (with a degree smaller than or equal to $\omega$) along with the incident edges can be safely removed from the graph. Our algorithm integrates this idea to dynamically prune the graph under consideration, which proves to be highly effective on massive sparse graphs. 

Finally, applying the pruning techniques can disconnect the original graph into several connected subgraphs. This observation can be explored advantageously to further prune the graph in combination with an exact algorithm. Indeed, if a subgraph is small enough such that an exact algorithm can identify the \textit{maximum} balanced biclique quickly, then the subgraph can be definitively removed since the subproblem (associated to the subgraph) is optimally solved. Moreover, the optimal solution of this subgraph can also be used to update the current best balanced biclique (and the lower bound bound), which can lead to additional reduction of the graph.

\subsection{General procedure}
\label{subsec_generalproc}

Based on the rationale presented in Section \ref{Rationale}, we introduce Tabu Search with Graph Reduction for MBBP (TSGR-MBBP) (Algorithm \ref{Algorithm_General}). TSGR-MBBP is an iterated two phase algorithm and includes two main components: the Constraint-Based Tabu Search (CBTS) procedure and the graph reducing procedure. The CBTS procedure is used to find high quality bicliques in the relaxed search space $\Omega^2$, while the graph reducing procedure aims to shrink progressively the current graph without losing optimal solutions.

After setting the best biclique $(X^b, Y^b)$ to $(\emptyset,\emptyset)$ and the best balanced size $\omega$ to 0 (lines 2 and 3), the algorithm repeats the main `while' loop (lines 4-20) until a stopping condition is met. For each `while' loop, an initial biclique, which is not necessarily balanced, is first generated by Random\_Init\_Solution() (line 5, see Section \ref{sec_random_initial}), and then further improved by the CBTS procedure (Constraint\_Tabu\_Improve(), line 6, see Section \ref{sec_tabu_search}). If the resulting biclique has a balanced size larger than the current best balanced size $\omega$, the best biclique $(X^b, Y^b)$ and the best balanced size are updated (lines 7-9). 

\begin{algorithm}[H]
\begin{small}
\caption{Main framework of TSGR-MBBP}
\label{Algorithm_General}
\KwIn{Graph instance $G=(U,V, E)$, tabu search depth $L$, cardinality threshold $K$ for graph reduction with exact algorithm, tabu tenure parameter $\alpha$.}
\KwOut{The maximum balanced biclique.}
\Begin{
$(X^b, Y^b) \gets (\emptyset,\emptyset)$\tcc*[r]{The largest balanced biclique found so far}
$\omega = 0$ \tcc*[r]{The largest balanced size found so far}
\While{stopping condition is not met}{
	\tcp{Find an improved biclique from a new initial biclique}		
	$(X,Y) \gets Random\_Init\_Solution(G)$	\tcc*{Section \ref{sec_random_initial}}
	$(X,Y) \gets Constraint\_Tabu\_Improve(G, (X,Y), L, \alpha)$ \tcc*{Section \ref{sec_tabu_search}}
	\If{$min(|X|,|Y|) > \omega$}{
		$(X^b,Y^b) \gets (X,Y)$\;
		$\omega \gets min(|X|,|Y|)$
	}
		\tcp{Graph reduction procedure using improved balanced size $\omega$}		
	\While{$\omega \geq min_{v\in U\cup V}\{|N(v)|\}$}{		
		\tcp{The first graph reduction}		
		$G \gets Peel(G, \omega)$ \tcc*{Section \ref{subsec_peel}}
		\tcp{The second graph reduction}				
		\For{each connected subgraph $G_i[U_i\cup V_i]$ in $G$}{
			\If{$|U_i|+|V_i| \leq K$}{
				$(X,Y) \gets Exact\_Search(G_i, \omega)$ \tcc*{Section \ref{subsec_exact_search}}
				\If{$min(|X|,|Y|) > \omega$}{ 
					$(X^b,Y^b) \gets (X,Y)$\;
					$\omega \gets min(|X|,|Y|)$
				}
				$G \gets G[(U\setminus U_i)\cup (V\setminus V_i)]$	
			}
		}
	}
	\If{$|U| \leq \omega \lor |V| \leq \omega$}{
		\Return{Make\_Balance$(X^b, Y^b)$}	\tcc*[r]{$(X^b, Y^b)$ is an optimum solution}
	}
} 
} 
\Return{Make\_Balance($X^b, Y^b$)}
\end{small}
\end{algorithm}

Now, if the current best balanced size is greater than or equal to the degree of any vertex in the current graph, the graph reduction procedure is activated (lines 10-18). This procedure includes two phases: first, reducing the current graph by the $Peel$ procedure to remove fruitless vertices and their incident edges (line 11, see Section \ref{subsec_peel}); second, determining the maximum balanced size of each connected subgraphs with up to $K$ (a predefined parameter) vertices by a branch-and-bound (B\&B) exact algorithm (Exact\_Search(), line 14, see Section \ref{subsec_exact_search}) and then deleting these subgraphs from the current graph (line 18). The optimal solution found by exact search can also be used to update the current best solution found so far (lines 15-17). Finally, though TSGR-MBBP is a heuristic algorithm, thanks to the graph reduction procedure, $\omega$ is proven to be the optimal balanced size when the cardinality of any partition ($|U|$ or $|V|$, which is a upper bound of the maximum biclique) in the current graph is no more than $\omega$ (which is a lower bound) (lines 19-20).

As explained in Section \ref{sec_prob_formul}, the proposed algorithm operates on the relaxed biclique space $\Omega^2$. As a result, the current solution $(X,Y)$ and the best biclique found so far $(X^b, Y^b)$ are not necessarily balanced with nevertheless a balance deviation limited to 2. Actually, the three procedures: Random\_Init\_Solution(), Constraint\_Tabu\_Improve() and Exact\_Search() generate or return a biclique with a balance deviation no more than 2. The procedure of retrieving a strict balanced biclique of size $\omega$ from an unbalanced biclique is accomplished by Make\_Balance(). This procedure simply removes vertices from the larger set $X^b$ or $Y^b$ until a balanced biclique of size $\omega$ is obtained. Obviously, no more than 2 vertices will be removed by Make\_Balance().

\subsection{Construct random initial solutions}
\label{sec_random_initial}

The Random\_Init\_Solution() procedure is invoked to initialize each restart of TSGR-MBBP with a new biclique. This procedure starts from a trivial solution formed by a random vertex from $U\cup V$, say $(X,Y)=(\{1\},\emptyset)$ (without loss of generality). Then, it iteratively expands the current solution by alternatively adding one vertex $v$ to the set $X$ or $Y$, $v$ being necessarily connected to all vertices of the other set. Specifically, in the first iteration, a vertex is selected randomly from the candidate set $\cap_{i\in X}N(i)\setminus Y$. Then, in the next iteration, we switch to the candidate set $\cap_{i\in Y}N(i)\setminus X$. The procedure continues until the current candidate set becomes empty. The time complexity of this procedure is bounded by $O(|U\cup V|\times|E|)$.

Consider Figure \ref{fig_demo1} as an example and suppose that we start from solution $(X,Y)=(\{1\},\emptyset)$, the algorithm expands the solution by selecting an arbitrary vertex from $N(1)\setminus \emptyset = \{5,6,8\}$ (say $5$) in the first iteration. In the second iteration, the algorithm expands $Y$ by adding a vertex from $N(5)\setminus\{1\} = \{2,3,4\}$. Suppose that the algorithm goes on likewise to achieve a solution $(X,Y)=(\{1,2,3\},\{5,6\})$ after four iterations. Then in the fifth iteration, we try to expand $Y$ by adding a vertex from the candidate set $\cap_{i \in N(X)}\setminus Y$. However, since this candidate becomes empty, the Random\_Init\_Solution() procedure stops and returns $(X,Y)=(\{1,2,3\},\{5,6\})$ as its output.

The biclique $(X,Y)$ returned by this procedure may not be strictly balanced, but the balance deviation can never exceed 1. This biclique is served as the starting solution for the tabu search procedure which is explained below. 

\subsection{Constraint-Based Tabu Search}
\label{sec_tabu_search}

The CBTS procedure (Algorithm \ref{algorithm_tabu_search}) is the main search component of the proposed algorithm. CBTS iteratively transforms the current solution (biclique) to a neighbor solution while respecting the unbalance limit of 2. The parameter $L$ (a positive integer) is called tabu search depth, which defines the total number of iterations of tabu search. The other parameter, $\alpha \in \mathbb{R}_+\cup\{0\}$, is a coefficient of tabu tenure (see Section \ref{subsubse_explore_ns}). In each iteration, CBTS applies the ``push" operator (lines 5-14, see Section \ref{subsubsec_push}), which either adds a vertex to the current solution or swaps a vertex of the biclique against a vertex outside of the biclique. Whenever the balance deviation exceeds 2 after an application of ``push", a repairing procedure is followed to recover the balance of the current biclique (lines 15-23, see Section \ref{subsubse_explore_ns}). The repairing procedure simply drops vertices from the larger partition of the biclique until the cardinality of both partitions becomes equal. CBTS terminates after $L$ such ``push" and ``repair" iterations.

\subsubsection{The push operator}
\label{subsubsec_push}

The ``push'' operator was first proposed for the maximum weight clique problem in \citep{Zhouetal2017} where each application of ``push'' adds a vertex (taken from a candidate set) in the clique and expels $p\geq 0$ vertices from the clique to maintain the feasibility of the transformed clique. In the context of MBBP, given a biclique $(X,Y)$ with $X\subseteq U$ and $Y\subseteq V$, and without loss of generality, suppose that a vertex $v \in N(Y) \setminus X$ (i.e., $N(Y) \setminus X$ is the candidate set for ``push'') is chosen. The ``push'' operator adds vertex $v$ to $X$ and expels from $Y$ the vertices that are not adjacent to $v$. Let $(X',Y')$ denotes the new biclique after the ``push'' operation, then we represent this transformation by $(X',Y') \gets (X,Y) \oplus push(v)$.

Similarly, if $v\in N(X)\setminus Y$, $(X',Y') = (X\cap {N(v)}, Y\cup \{v\})$, $\delta_v$ is updated by the same rule except that the roles of $X$ and $Y$ are exchanged. 

The ``push" operator can be explained as a generalization of the conventional $(1,p)$-swap ($p\in Z^0$) operator. For example, if we restrict the candidate vertex $v$ with property $N(v)\cap X = X$ or $N(v)\cap Y = Y$,  $push(v)$ is equivalent to adding $v$ without expelling any vertex (i.e., $(1,0)$-swap); if we restrict $v$ with property $|X\setminus  N(v)| = 1$ or $|Y\setminus N(v)| = 1$, $push(v)$ exchanges $v$ with another vertex in $X$ or $Y$ that is not adjacent to $v$ (i.e., $(1,1)$-swap). Actually, the two restrictions are employed in our CBTS algorithm to customize the ``push" operator, as explained in the next section.

Let $\delta_v = min(|X'|,|Y'|) - min(|X|,|Y|)$ be the change of the balanced sizes between $(X',Y')$ and $(X,Y)$, then  $\delta_v$ can be calculated by the following rule.

\begin{equation}
\label{equation_delta}
\delta_v \leftarrow 
\left\{
\begin{array}{ll}
-|Y\setminus N(v)| & {\rm, if \ } |X|>|Y|\\
min(1, |Y|-|X|-|Y\setminus N(v)|) & {\rm , otherwise}
\end{array}
\right.
\end{equation}

\begin{algorithm}[H]
\begin{small}
\caption{Constraint-Based Tabu Search}
\label{algorithm_tabu_search}
\KwIn{Graph instance $G=(U,V,E)$, starting solution $(X,Y)$, tabu search depth $L$, tabu tenure parameter $\alpha$.}
\KwOut{The best biclique $(X^*, Y^*)$ found.}
\Begin{
$I \gets 0$, $(X^*, Y^*) \gets (X,Y)$\tcc*[r]{$I$ is the iteration counter, $(X^*, Y^*)$ keeps the best biclique found so far}
$T[1...n]\gets [0...0]_{n}$ \tcc*[l]{initiate tabu list, each vertex $v$ being marked tabu for the next $T[v]$th iterations; $n =|U|+|V|$}
\While{$I \leq L$}{
	\tcp{Explore the neighbor solutions}
	Build $C_{expand} \subseteq C$ and $C_{plateau} \subseteq C$ \tcc*[r]{Decompose candidate set, see Section \ref{subsubse_explore_ns}}
	$v \gets null$\;
	\If{$C_{expand} \neq \emptyset$}{
		$v\gets random(C_{expand})$ \;
	}\ElseIf {$C_{plateau}\neq \emptyset$}{
		$v\gets random(C_{plateau})$ \;		
	}
	\If {$v \neq null$}{
		$(X,Y) \gets (X,Y) \oplus push(v)$ \;
		\tcp{Set tabu tenure for each vertex expelled by push.}
		\For{$u \gets $ expelled vertex}{ 
			$T[u] \gets I + tt(\alpha, |S|)$ \tcc*[r]{$S = X$ if $u\in X$, otherwise $S = Y$}
		}		
	}
	\tcp{Recover balance when the balance deviation  exceeds 2}
	\If{$||X|-|Y|| > 2$}{
		\While {$|X| > |Y|$}{
			$u \gets random(X)$ 	\;
			$(X,Y) \gets (X,Y) \oplus drop(u)$ \;
			$T[u] \gets I + tt(\alpha, |X|)$ {\tcc*[r]{Set tabu tenure for the dropped vertex}
		}}
		\While {$|X| < |Y|$}{
			$u \gets random(Y)$ \;
			$(X,Y) \gets (X, Y) \oplus drop(u)$ \;
			$T[u] \gets I + tt(\alpha, |Y|)$	 \;
		}
	}
	\tcp{update the best solution}
	\If{$min(|X|,|Y|) > min(|X^*|,|Y^*|)$}{
		$(X^*, Y^*) \gets (X,Y)$
	}
	$I \gets I + 1$
} 
} 
\Return{$(X^*, Y^*)$}
\end{small}
\end{algorithm}

\subsubsection{Explore the neighbor solutions}
\label{subsubse_explore_ns}

The general ``push" operator applied to MBBP can add an arbitrary vertex from the candidate set $N(X\cup Y)$ into one set $X$ or $Y$, and then expel $p \geq 0$ vertices from the other set. However, for the reason of computational efficiency, only a subset of $N(X\cup Y)$ is considered for each ``push" operation. Specifically, we add restrictions on the candidate vertices for the ``push" operation so that it adds one vertex to the current solution and at the same time, no more than one vertex from the current solution will be expelled. These restrictions lead exactly to the two cases that were introduced at the end of Section \ref{subsubsec_push}. In Algorithm \ref{algorithm_tabu_search}, set $C$ (line 5) includes the restricted candidate vertices for ``push". Every vertex in $C$ is adjacent to all the vertices of $X$ (or $Y$), or all but one vertex of $X$ (or $Y$).

Moreover, $C_{expand}$ is a subset of $C$ such that applying ``push" to any vertex (say $v$) of this subset always results in a solution of better quality (i.e., $\delta_v > 0$). Similarly, $C_{plateau} \subseteq  C$ includes the vertices that can be exchanged by ``push" to obtain solutions of equal quality (i.e., $\delta_v = 0$).  

To prevent CBTS from revisiting recently examined solutions, a tabu list \citep{glover1997tabu} is considered when we construct $C_{expand}$ and $C_{plateau}$ from candidate set $C$: a vertex which is marked tabu in the current iteration will not be included in $C_{expand}$ or $C_{plateau}$ unless pushing the vertex into the solution leads to a solution better than the best solution ever found (this is called aspiration rule in tabu search terminology). To sum up, let $(X,Y)$ and $(X^*,Y^*)$ be respectively the current solution and  the best solution found so far during the current CBTS run, $I$ the current iteration number, $T[v]$ the tabu tenure of vertex $v$ (see below), then the restricted candidate set $C$, and sets $C_{expand}$, $C_{plateau}$ are defined as follows.

\begin{equation} 
\label{eq_set_defines}
\begin{split}
&C =\{v \in N(X\cup Y): v\in U \land |N(v)\cap Y|\geq |Y|-1, v\in V \land |N(v)\cap X|\geq |X|-1\}\\
&C_{expand} = \{v \in C: \delta_v > 1, T[v] \leq I \lor min(|X|,|Y|)+1>min(|X^*|,|Y^*|)\} \\
&C_{plateau} = \{v \in C: \delta_v = 0, T[v] \leq I\}
\end{split}
\end{equation}

where $T[v] \leq I$ indicates that vertex $v$ is no more forbidden by the tabu list for the current iteration and can take part in a future ``push" operation.

Given the subsets $C_{expand}$, $C_{plateau}$ as two alternative candidate sets for ``push", CBTS gives priority to  $C_{expand}$ since pushing vertices of this set always improves the current biclique. Only when $C_{expand}$ is empty, set $C_{plateau}$ is explored by the ``push" operator (lines 6-14). After each ``push" application with $C_{plateau}$, the vertex expelled  by ``push" ($u$ in line 13) is marked tabu for the following $tt(\alpha, |A|)$ ($A = X$ if $u\in X$, otherwise $A = Y$) consecutive iterations.
According to \citep{Zhouetal2017}, $tt(\alpha,l)$ (called tabu tenure) is defined by the function: $tt(\alpha, l) = max(7, \alpha * random(l))$ where $\alpha \in \mathbb{R}_+\cup\{0\}$ is a predefined parameter and $random(l)$ returns a random integer in $[0,l]$.


\subsubsection{Recover biclique balance}
\label{Recover biclique balance}

Recall that with the restrictions on candidate vertices, the number of vertices expelled by ``push'' in each iteration is either zero or one. As a result, if the balance deviation of the current biclique is greater than 2 (i.e., $||X|-|Y||>2$), it is impossible to make the biclique strictly balanced with one application of the ``push" operator. Consequently, each time the balance deviation of the solution exceeds 2, we restore the balance property by applying a repair procedure (lines 15-23). This repair procedure simply drops vertices from the larger set ($X$ or $Y$) of the biclique until the solution becomes strictly balanced (denoted as $(X,Y) \gets (X,Y) \oplus drop(u)$ at lines 18 and 22). Again, each dropped vertex $u$ is forbidden to rejoin the solution during the period fixed by its tabu tenure ($tt(\alpha, |X|)$ if $u\in U$, $tt(\alpha, |Y|)$ if $u\in V$). In general, CBTS utilizes the ``push'' operator to explore the space $\Omega^2$ rather than $\Omega^0$ by constraining the balance deviation of the visited solutions. In Section \ref{subsec_effect_cbts}, we further investigate the effectiveness of this strategy.

\subsubsection{Time complexity}
\label{Implementation}

CBTS operates directly on the input graph and uses the adjacent list representation to store the graph. Given a solution $(X,Y)$, by our implementation, the time complexity of constructing $C_{expand}$ and $C_{plateau}$ is bounded by $O(|N(X\cup Y)|)$. The time complexity of moving one vertex (outside the solution or into the solution) is bounded by $O(M)$ ($M=\max_{v\in U\cup V}\{|N(v)|\}$). Hence, the time complexity of one iteration in CBTS is bounded by $O(|N(X\cup Y)|+2\times M)$. Though $|N(X\cup Y)|$ is almost equal to $|U|+|V|$ in dense graphs, in very large real-life networks, both $|N(X\cup Y)|$ and $M$ are very limited due to the sparsity of the graphs.


\subsection{Reduction by the $Peel$ procedure}
\label{subsec_peel}

Our TSGR-MBBP algorithm employs the $Peel(G,w)$ procedure (Algorithm \ref{Algorithm_General}, line 11) to recursively delete all vertices whose degrees are smaller than or equal to $\omega$ until no such vertex exists. Obviously, if the cardinality of one vertex set of the reduced bipartite graph (which is a upper bound of the maximum biclique) is less than or equal to $\omega$ (which is a lower bound), then $\omega$ must be the optimal objective value because no better solution can exist in the reduced graph (Algorithm \ref{Algorithm_General}, lines 19-20).

The peeling procedure is triggered each time the balanced size of the largest biclique discovered so far (lower bound) is larger than or equal to the minimum degree of the current graph. This procedure is effective on large sparse graphs but may not reduce a dense graph much. The experiments reported in Section \ref{sec_experiments} confirm that, with a high quality lower bound, large real-life bipartite graphs can be significantly reduced.

We note that the idea of reducing a graph by removing unpromising vertices was previously used in a GRASP heuristic for detecting dense subgraphs (quasi-cliques) in massive sparse graphs \citep{abello2002massive}. We adapted this technique for solving MBBP for the first time.

\subsection{Reduction by exact search}
\label{subsec_exact_search}

Exact search algorithms guarantee the optimality of the solution found, but may require prohibitive computing time on large instances. However, since exact search algorithms are able to prove optimality on small graphs rapidly, they can still be used as a basis for  graph reduction. In Algorithm \ref{Algorithm_General} (lines 12-18), we show such an approach of using exact search for MBBP. If a solution has been confirmed to be optimal for a subgraph of the current graph, this subgraph can be safely eliminated from the current graph. Moreover, since the optimal value of the subgraph is a lower bound of the initial graph, we can use the optimal solution of the subgraph to update the current best balanced biclique, which in turn can further reduce the current graph. The exact algorithm used by TSGR-MBBP was adapted from a well-known B\&B algorithm for the maximum clique problem \citep{carraghan1990exact} and described in Appendix \ref{apx_sec_exact_search}. This exact algorithm is only applied to solve a subgraph with $K$ vertices at most ($K$ being the largest subgraph that is estimated to be solved in reasonable time by the algorithm). It is clear that $K$ depends on the adopted exact algorithm and target subgraph. According to our experiments, we set $K$ to 100 for random dense graphs and 500 for sparse real-life networks.

\section{Computational assessment}
\label{sec_experiments}

To comprehensively evaluate the proposed TSGR-MBBP algorithm as well as its components, we tested our algorithm on two sets of benchmark instances including both (dense) random graphs and massive real-life networks.

\subsection{Benchmark}
\label{subsec_benchmark}

\begin{itemize}
\item Random Graphs: This set of benchmark instances includes 30 randomly generated dense graphs. In each graph, the two vertex sets $U$ and $V$ have  an equal cardinality (i.e., $|U|=|V|$) and an edge between a pair of vertices $(u,v) \in U\times V$ exists with uniform probability $p$ ($0<p<1$) which defines the edge density of the graph. For our study, we used random graphs generated by the same rule of \citep{yuan2015new} so that the performances of different algorithms can be compared. For each combination of $n \in \{250,500\}$ and $p \in \{0.85,0.90,0.95\}$ ($n=|U|=|V|$), 5 instances were generated (30 in total). These instances are thus very dense and named as ``G\_$<$n$>$\_$<$p$>$\_$<$id$>$" where $id \in \{1,2,3,4,5\}$. A theoretical analysis in \citep{dawande2001bipartite} showed that the maximum balanced size $\omega$ in random graphs locates in range $[\frac{\ln n}{\ln(1/p)},\frac{2* \ln n}{\ln(1/p)}]$ with high probability (when $n$ is sufficiently large).
\item The Koblenz Network Collection (KONECT) \citep{kunegis2013konect}: The entire collection contains hundreds of networks derived from different real-life applications, including social networks, hyperlink networks, authorship networks, physical networks, interaction networks and communication networks. Though  KONECT dataset was originally designed for network analysis, these large bipartite networks are also suitable for testing TSGR-MBBP. We used 25 bipartite graphs varying from smaller ones (829 + 551 vertices and 1476 edges) to very large ones (1,425,813 + 4,000,150 vertices and 8,649,016 edges). Irrelevant graph information for MBBP like multiple edges, vertex or edge weight in some graphs is ignored.
\end{itemize}

\subsection{Parameter tuning and experimental protocol}
\label{subsec_exp_protocoal}

The TSGR-MBBP algorithm has three parameters: $L$ - the tabu search depth; $\alpha$ - the coefficient for tabu tenure required by the Constraint\_Tabu\_Improve() procedure (Section \ref{sec_tabu_search}); $K$ - the threshold on the number of vertices of the subgraph for graph reduction with the exact algorithm (Section \ref{subsec_exact_search}).

Since the first two parameters ($L$ and $\alpha$) are independent from the reduction procedure, we tuned them on a simplified version of TSGR-MBBP without the graph reduction procedure (i.e., by disabling lines 10-20 in Algorithm \ref{Algorithm_General}). We used the automatic parameter configuration package iRace \citep{lopez2011irace}, which implements the Iterated F-Race (IFR) method. Given $L \in \{10,100,1000,5000,10000\}$, and $\alpha \in [0,2]$, for each parameter configuration, we used a tuning budget of 500 hook-runs, each of which representing 10 independent calls of TSGR-MBBP. The training set for random graphs included 6 challenging instances, i.e., GraphU\_500\_XXX\_1.clq and GraphU\_500\_XXX\_2.clq (XXX can be replaced by 0.95, 0.90, 0.85). The experiments suggested that the combination ($L=1000, \alpha = 0.30$) was a suitable configuration for random graphs. As for KONECT graphs, the training set included ``actor-movie", ``bookcrossing\_full-rating", ``dbpedia-genre", ``dbpedia-team", ``github", ``stackexchange-stackoverflow". The final choice of parameters was $L=100$ and $\alpha=1.74$.

The use of two different settings for $(L,\alpha)$ is mainly due to the graph structures which vary much. According to our observations, for random dense graphs, a more intensified search is needed to find quality solutions. This is achieved with a large tabu search depth ($L=1000$) and a short tabu tenure (with $\alpha = 0.30$). On the contrary, for large real-life sparse instances, the tabu search component is able to reach local optima very quickly. As a result, it is preferable to restart more frequently the tabu search component (with $L=100$) and diversify more strongly the search during the optimization process (using a larger tabu tenure with $\alpha=1.74$).

The third parameter $K$ indicates the largest subgraph that can be solved in reasonable time by the exact algorithm described in Section \ref{subsec_exact_search}. We set $K=100$ for random graphs and $500$ for KONECT graphs. In effect, since the random graphs we tested are very dense, they cannot be reduced by the reduction procedure, implying that no connected subgraph with less than 100 vertices exists in this set of benchmarks. A very large $K$ is not acceptable, otherwise the computing time for exact search becomes prohibitive according to our observations for random graphs. Preliminary experiments also confirmed that the time consumption was normally insignificant (less than 2 seconds) for connected subgraphs with less than 500 vertices for sparse KONECT graphs. As the vertex number is just a rough estimation of the hardness of the subgraph for our exact algorithm, we terminate the exact algorithm if it does not finish during 10 seconds. This additional cutting-off condition prevents the algorithm from spending too much effort in searching optimal solutions for some potential hard subgraphs. If the exact search stops without giving an optimal solution, the corresponding subgraph will not be removed. 

TSGR-MBBP was implemented in C++ and compiled with g++ v4.4.7 with optimization flag -o3. Our experiments were performed on a computer with an AMD Opteron 4184 processor (2.8GHz and 2GB RAM) running Linux 2.6.32. When solving the DIMACS machine benchmark procedure `dfmax.c'\footnote{\url{dfmax: ftp://dimacs.rutgers.edu/pub/dsj/clique/}} without compilation optimization flag, the run time on our machine is 0.40, 2.50 and 9.55 seconds for graphs r300.5, r400.5 and r500.5 respectively.

Considering the stochastic nature of TSGR-MBBP, we ran TSGR-MBBP 20 independent times to solve each instance. For the random graphs of 250 vertices, the time limit of each run was 30 seconds, while for the random graphs of 500 vertices, 60 seconds were allowed. As for the KONECT instances, we prolong this limitation to 360 seconds (6 minutes) since these instances are much larger than the random graphs.

\subsection{Computational results}

\subsubsection{Random graphs}
\label{subsec_random_graph}

To evaluate the performance of TSGR-MBBP, we show computational results relative to three state-of-the-art MBBP approaches:

\begin{itemize}
\item[-] EA/SM \citep{yuan2015new}: This is a hybrid algorithm mixing local search, structure mutation and repair-assisted restart. EA/SM is the most recent heuristic algorithm and outperforms the precedent algorithms like in \citep{yuan2011low,yuan2014fast}. For our comparative experiment, we ran 20 times the source code of EA/SM (provided by its authors) to solve each instance, each run being limited to 200,000 fitness evaluations according to \citep{yuan2015new}. We observed that to attain its best solutions, EA/SM needed a run time ranging from 42 to 50 seconds for instances of 250 vertices and 75 to 94 seconds for instances of 500 vertices (see Table \ref{tbl_random_ins}). Consequently, the stopping condition of EA/SM can be considered to be more favorable than that used to run our algorithm (a \textit{cut off time} of 30 seconds for instances of 250 vertices and and 60 seconds for instances of 500 vertices).

\item[-] IBM CPLEX: CPLEX is one of the most popular commercial optimization software. We ran CPLEX (version 12.6.1) 2 hours (7200 seconds) on each instance with the binary linear formulation provided in Section \ref{sec_introduction}. Obviously, the total time given to TSGR-MBBP for 20 runs (60*20 = 1200 seconds for the random instances and 360*20 = 7200 seconds for the KONECT instances) is no more than 2 hours.

\item[-] AL\_Greedy \citep{al2007defect}. This is a (fast) greedy algorithm which solves the equivalent maximum balanced independent set problem for the bipartite complement. According to \citep{yuan2015new}, this algorithm performs better than its earlier version presented in \citep{tahoori2006application}. Thus, we re-implemented this algorithm and used it for our comparative study. Since AL\_Greedy is a deterministic heuristic, only one run was needed to solve each instance. Moreover, AL\_Greedy stops once its construction procedure reaches its end. Thus, no explicit stopping condition is required.
\end{itemize}

	\begin{table}\centering	
	\small
	\begin{scriptsize}
	\caption{Computational results of TSGR-MBBP together with the results of EA/SM \citep{yuan2015new}, CPLEX (version 12.6.1) and AL\_Greedy \citep{al2007defect} on the set of 30 random dense graphs. }	
	\label{tbl_random_ins}
	\begin{tabular}{c|c|ccc|cc|cc|c}
	\hline
	\multirow{2}{*}{instance}&\multirow{2}{*}{BKV} & \multicolumn{3}{c|}{TSGR-MBBP} & \multicolumn{2}{c|}{EA/SM}  & \multicolumn{2}{c|}{CPLEX 12.6.1}  &AL\_Greedy\\
	\cline{3-10}
	   & & best(ave) & time & reduce & best(ave) & time & best & time & best\\
	\hline 
G\_250\_0.95\_1  &68 &68 &0.05 &0 &68(67.90) &50.28 &66 &$\geq$7200 &64 \\
G\_250\_0.95\_2  &66 &66 &0.21 &0 &66(65.05) &49.31  &64 &$\geq$7200 &59 \\
G\_250\_0.95\_3  &70 &70 &0.17 &0 &70(69.50) &48.87  &- &- &67\\
G\_250\_0.95\_4  &68 &68 &0.42 &0 &68(67.10) &47.36   &66 &$\geq$7200 &63\\
G\_250\_0.95\_5  &68 &68 &0.72 &0 &67(66.95) &47.41  &67 &$\geq$7200 &62\\
G\_250\_0.90\_1  &44 &44 &0.06 &0 &44(43.70) &42.94  &42 &$\geq$7200 &37\\
G\_250\_0.90\_2  &44 &\textbf{45} &0.52 &0 &45(43.90) &43.28  &42 &$\geq$7200 &39\\
G\_250\_0.90\_3  &44 &44 &0.13 &0 &44(43.45) &43.20  &42 &$\geq7200$ &40\\
G\_250\_0.90\_4  &45 &45 &0.66 &0 &44(43.80) &43.13  &42 &$\geq7200$ &40\\
G\_250\_0.90\_5  &45 &45 &0.23 &0  &45(44.10) &45.13  &41 &$\geq$7200 &40\\
G\_250\_0.85\_1  &33 &33 &0.11 &0 &33(32.40) &47.92  &- &- &30\\
G\_250\_0.85\_2  &33 &33 &0.04 &0 &33(32.75) &49.94  &- &- &31\\
G\_250\_0.85\_3  &34 &34 &0.69 &0 &34(32.95) &44.66  &- &- &31\\
G\_250\_0.85\_4  &33 &33 &0.07 &0 &33(32.90) &43.76  &30 &$\geq$7200 &30\\
G\_250\_0.85\_5  &33 &33 &0.52 &0 &33(32.30) &44.16  &30 &$\geq$7200 &30\\
G\_500\_0.95\_1  &91 &\textbf{93} &14.37 &0 &91(90.20) &93.28  &- &- &83\\
G\_500\_0.95\_2  &89 &\textbf{91} &15.58 &0 &90(88.30) &92.02  &- &- &81\\
G\_500\_0.95\_3  &89 &\textbf{91}(90.05) &3.85 &0 &90(87.85) &92.62  &85 &$\geq7200$ &81\\
G\_500\_0.95\_4  &88 &\textbf{90}(89.40) &21.04 &0 &88(86.85) &93.28  &83 &$\geq7200$ &78\\
G\_500\_0.95\_5  &90 &\textbf{91}(90.90) &13.40 &0 &90(88.15) &94.30  &81 &$\geq7200$ &83\\
G\_500\_0.90\_1  &56 &56 &12.21 &0 &55(53.75) &76.24  &46 &$\geq7200$ &49\\
G\_500\_0.90\_2  &56 &56 &5.38 &0 &56(54.00) &79.34  &47 &$\geq7200$ &48\\
G\_500\_0.90\_3  &54 &\textbf{56}(55.60) &15.57 &0 &55(53.45) &79.52  &46 &$\geq7200$ &48\\
G\_500\_0.90\_4  &55 &\textbf{56}(55.55) &9.87 &0 &55(53.75) &79.59  &47 &$\geq7200$ &48\\
G\_500\_0.90\_5  &55 &\textbf{56}(55.50) &13.68 &0 &55(53.25) &82.23  &44 &$\geq7200$ &48\\
G\_500\_0.85\_1  &40 &40 &4.59 &0 &40(38.45) &75.55  &33 &$\geq7200$ &34\\
G\_500\_0.85\_2  &41 &41 &5.84 &0 &40(39.25) &75.56  &32 &$\geq$7200 &33\\
G\_500\_0.85\_3  &40 &\textbf{41}(40.50) &13.50 &0 &41(38.65) &81.48  &35 &$\geq$7200 &35\\
G\_500\_0.85\_4  &40 &40 &1.84 &0 &39(38.30) &75.29  &33 &$\geq$7200 &35\\
G\_500\_0.85\_5  &41 &41 &4.60 &0 &40(38.60) &75.06  &31 &$\geq$7200 &34\\
	\hline
	\end{tabular}
	\end{scriptsize}
	\end{table}	
	
Table \ref{tbl_random_ins} reports the computational results of TSGR-MBBP together with the results of the reference approaches (EA/SM, CPLEX and AL\_Greedy) on the 30 random dense graphs. Column ``instance'' shows the name of each instance. Column ``BKV" presents the best known values reported in \citep{yuan2015new}. For TSGR-MBBP and EA/SM, column ``best(ave)" indicates the maximum value of the 20 best balanced sizes found in 20 runs, the average size is given between parentheses if the 20 runs do not lead to an identical balanced size; column ``time'' reports the average time (in seconds) of first hitting the best balanced size in 20 runs; column ``reduce'' (only for TSGR-MBBP) reports the number of vertices removed by the two reduction methods in one of the runs where we find the best balanced size. For CPLEX, we report the best lower bounds and the time needed to complete the search. If CPLEX fails to report a feasible solution for an instance due to memory limitation, ``-" is used in the corresponding entries of columns ``best'' and ``time''. For AL\_Greedy, since its run time is negligible (shorter than 0.01 second for all instances), we only report the best biclique values.

From Table \ref{tbl_random_ins}, we first observe that in terms of solution quality, TSGR-MBBP competes very favorably with the reference approaches. In particular, TSGR-MBBP improves the best-known results reported in \citep{yuan2015new} for 10 instances (marked in bold font). For the 20 remaining instances, the best objective values found by TSGR-MBBP are always as good as or better than those of the reference algorithms. The average objective values of the 20 runs of TSGR-MBBP are also better than that of EA/SM. Moreover, the performance of TSGR-MBBP is quite stable across the whole set of tested instances. In terms of  computational efficiency, TSGR-MBBP is very competitive -- it hits its best result within no more than one and 22 seconds for the instances of 250 and 500 vertices respectively, against up to 50 and 94 seconds for the best reference algorithm EA/SM. As for CPLEX, it cannot complete its search within a duration of 2 hours and thus fails to find the optimal solution for any instance (still CPLEX finds some solutions better than those of AL\_Greedy). Unsurprisingly, the greedy algorithm AL\_Greedy leads to solutions of very poor quality. Finally, as expected, neither reduction method is successful on these very dense graphs as the degree of any vertex is much larger than the best balanced size. For example, the vertex degree of ``G\_500\_0.85\_X'' is closely around 425 while the optimal balanced size is estimated to be between 39 and 76 by the theorem given in \citep{dawande2001bipartite}. However, as we show in the next section, the reduction procedure becomes extremely effective when large sparse graphs are considered. 

\subsubsection{KONECT networks}
\label{subsec_exp_konect}

	\begin{table}\centering	
	\small
	\begin{scriptsize}
	\caption{Computational results of TSGR-MBBP and CPLEX on the set of 25 large KONECT instances. The results of EA/SM and AL\_Greedy are not available.}
	\label{tbl_large_ins}
	\begin{tabular}{p{1.8cm}cp{1.0cm}|p{0.8cm}p{0.6cm}p{0.8cm}p{0.5cm}|p{2.0cm}p{0.5cm}p{0.5cm}}
	\hline
     \multicolumn{3}{c|}{instance}& \multicolumn{4}{c|}{TSGR-MBBP} & \multicolumn{3}{c}{CPLEX}\\
    	\hline
	name & $(|U|,|V|)$ & $|E|$ & best(ave) & time & red\_1 & red\_2 & $(|U'|,|V'|)$ &best & times\\
	\hline
actor-movie  &(127823, 383640) &1470418 &8 &8.91 &474822  &357  &(100398, 88729) &N/A &N/A\\
bibsonomy-2ui  &(5794, 767447) &2555080 &$8^*$ &1.01 &772062  &1179   &(137, 307) &8$^*$ &2209.76 \\
bookcrossing\_full-rating\   &(105278, 340523) &1149739 &13(12.30) &122.25 &433428  &33 &(26799, 76949) &N/A &N/A\\
dblp-author  &(1425813, 4000150) &8649016 &$10^*$ &8.92 &5416361  &9602 &(0, 0) &- &- \\
dbpedia-genre  &(258934, 7783) &463497 &$7^*$ &1.22 &265973  &744  &(385, 118) &7$^*$ &931.59 \\
dbpedia-location   &(172091, 53407) &293697 &$5^*$ &0.22 &224220  &1278  &(0, 0) &- &- \\
dbpedia-occupation &(127577, 101730) &250945 &$6^*$ &0.88 &228847  &460  &(0, 0) &- &- \\
dbpedia-producer  &(48833, 138844) &207268 &$6^*$ &0.17 &183879  &3798  &(0, 0) &- &- \\
dbpedia-recordlabel   &(168337, 18421) &233286 &$6^*$ &0.23 &186474  &284  &(0, 0) &- &- \\
dbpedia-starring &(76099, 81085) &281396 &$6^*$ &0.29 &156370  &814  &(44, 21) &6$^*$ &2.06 \\
dbpedia-team  &(901166, 34461) &1366466 &6(5.50) &99.29 &906083  &341  &(24858, 4345) &N/A &N/A \\
dbpedia-writer  &(89356, 46213) &144340 &$6^*$ &0.13 &131338  &4231  &(0, 0) &- &- \\
discogs\_affiliation   &(1754823, 270771) &14414659 &26 &22.15 &2008903  &662  &(11722, 4307) &N/A &N/A\\
discogs\_lgenre &(270771, 15) &4147665 &$15^*$ &10.16 &270786  &0 &(0, 0) &- &- \\
discogs\_style   &(1617943, 383) &24085580 &38(37.15) &131.85 &1612732  &0  &(5289, 305) &36 &$\geq 7200$ \\
edit-frwiki  &(288275, 4022276) &46168355 &41(27.50) &228.91 &4250247  &0   &(6664, 56700) &N/A &N/A\\
edit-frwiktionary  &(5017, 1907247) &7399298 &19 &31.88 &1909273  &0   &(232, 2759) &16 &$\geq 7200$\\
flickr-groupmemberships  &(395979, 103631) &8545307 &67 &94.74 &458053  &0 &(213863, 61790) &N/A &N/A\\
github  &(56519, 120867) &440237 &12 &4.74 &169775  &774   &(4001, 2836) &N/A &N/A\\
moreno\_crime &(829, 551) &1476 &$2^*$ &0.00 &1072  &308  &(4, 4) &2$^*$ &0.03\\
opsahl-collaboration &(16726, 22015) &58595 &$8^*$ &0.05 &37780  &961 &(0, 0) &- &- \\
opsahl-ucforum  &(899, 522) &33720 &$5^*$  &0.03 &531  &890  &(0, 0) &- &- \\
stackexchange-stackoverflow  &(545196, 96680) &1301942 &9(8.95) &92.12 &625399  &52  &(1432, 867) &8 &$\geq 7200$\\
wiki-en-cat  &(1853493, 182947) &3795796 &$14^*$ &17.58 &2027887  &8553  &(87, 60) &14$^*$ &11.23 \\
youtube-groupmemberships  &(94238, 30087) &293360 &12 &0.81 &121908  &118   &(1432, 867) &8 &$\geq 7200$ \\	
\hline
	\end{tabular}
	\end{scriptsize}
	\end{table}	
	
We report in Table \ref{tbl_large_ins} the computational results of TSGR-MBBP and CPLEX on the set of 25 KONECT instances. For this study, we ignore EA/SM and AL\_Greedy since the EA/SM code cannot be run on these graphs (EA/SM imposes the input graph to be balanced, which is not the case for KONECT instances), while AL\_Greedy performs very poorly (see Table \ref{tbl_random_ins}). Columns ``name'', ``$(|U|,|V|)$'', ``$|E|$'' show the basic information of the original instances. For TSGR-MBBP, columns ``best(ave)" and ``time" report the same information as in Table \ref{tbl_random_ins}. Columns ``red\_1'' and ``red\_2'' indicate the total number of vertices that are removed from the original graph by the two reduction methods (the $Peel$ procedure and the exact search procedure) in one of the runs where we find the best balanced size. To enable CPLEX to load large graphs, each original graph was pre-reduced by applying $Peel(G,best)$ before starting CPLEX. Column ``$(|U'|,|V'|)$" reports the number of vertices after applying $Peel(G,best)$ while columns ``best'' and ``time" report the best balanced size reached as well as the total consumed time. Symbol ``*" indicates that the solution has been proven to be optimal by the corresponding algorithm, while symbol ``-" means that the initial (and $Peel$ pre-reduced) graph cannot be loaded into CPLEX.

As explained in Section \ref{subsec_generalproc}, when either of the two vertex sets of the current bipartite graph contains less than $\omega$ (the best balanced size found so far) vertices, $\omega$ is proven to be the optimal maximum balanced size. From Table \ref{tbl_large_ins}, we observe that TSGR-MBBP proves optimality for 14 out of the 25 instances (indicated by ``*"), even though these real-world instances are significantly larger than the random instances. Also, TSGR-MBBP achieves the same best balanced size in all 20 runs for all but 5 instances (whose average objective values are reported in the table). Observing the number of vertices that has been reduced, we find that the first reduction method (the $Peel$ method) prunes more than half or even all of the vertices during the search procedure. As for the second reduction method (which is based on exact search), though the vertices removed by this method are fewer than the first method,  we cannot neglect its significance. For 5 instances ``bibsonomy-2ui",``dpedia-genre",``dbpedia-starring", ``moreno\_crime" and ``wiki-en-cat", the $Peel$ procedure fails to reduce these graphs to small enough  subgraphs such that optimality can be proven (one vertex set of the subgraph includes fewer than $\omega$ vertices, see column ``$(|U'|,|V'|)$"), TSGR-MBBP directly finds the optimal solution for the resulting subgraphs with less than $K$ vertices. The CPLEX solver, unfortunately, is unable to load some of these massive graphs even after reducing these graphs significantly by applying $Peel(G,best)$ in the pre-processing step. For instances for which CPLEX finds a feasible solution, like  ``discogs\_style", ``edit-frwiktionary",  ``stackexchange-stackoverflow" and ``youtube-groupmemberships", the results are still worse than those achieved by TSGR-MBBP. Besides, CPLEX always requires a longer time than TSGR-MBBP to attain the best solution.

\section{Analysis}
\label{sec_analysis}

This section presents an empirical analysis of the restricted unbalance constraint related to the Constraint-Based Tabu Search procedure (Section \ref{sec_tabu_search}) and the merit of the graph reduction procedure (Section \ref{subsec_peel}).

\subsection{Unbalance constraint of Constraint-Based Tabu Search}
\label{subsec_effect_cbts}

	\begin{table}\centering	
	\small
	\begin{scriptsize}
	\caption{Comparison between three different versions of the Constraint-Based Tabu Search procedure: CBTS$_{\Omega^\infty}$ can visit any biclique; CBTS$_{\Omega^1}$ visits only bicliques with a balance deviation no more than 1; CBTS$_{\Omega^2}$ (the original version of CBTS) visits bicliques with a balance deviation no more than 2. }
	\label{tbl_banalce_cmp}
	\begin{tabular}{c|ccc|ccc|ccc}
	\hline
    \multirow{2}{*}{instance} & \multicolumn{3}{c|}{CBTS$_{\Omega^\infty}$} & \multicolumn{3}{c|}{CBTS$_{\Omega^1}$} &\multicolumn{3}{c}{CBTS$_{\Omega^2}$}\\
    \cline{2-10}
       &best(ave) & time & iter &best(ave) & time & iter &best(ave) & time & iter \\
     \hline
GraphU\_500\_0.05\_3    &90(89.20) &8.48 &2827917 &70(68.25) &19.63 &1387704  &91(90.20) &7.76 &2297846 \\
GraphU\_500\_0.05\_4    &90(88.20) &19.48 &2886301&69(67.35) &19.47 &1386318  &90(89.20) &10.99 &2288732 \\
GraphU\_500\_0.05\_5    &91(89.95) &20.37 &2829679 &71(68.40) &28.71 &1389515  &91(90.95) &16.41 &2287010 \\
GraphU\_500\_0.10\_3    &54(53.85) &18.77 &5620994 &42(40.70) &15.32 &1387011 &56(55.60) &13.49 &2343998 \\
GraphU\_500\_0.10\_4   &55(54.20) &17.11 &5774901 &42(40.90) &16.74 &1385316  &56(55.30) &7.17 &2356738 \\
GraphU\_500\_0.10\_5    &55(54.10) &10.04 &5750204 &42(41.45) &22.12 &1388886 &56(55.55) &13.27 &2352404 \\
GraphU\_500\_0.15\_3    &39(38.55) &15.57 &6340722 &31(29.75) &18.62 &1441000 &41(40.70) &16.92 &2409855 \\
dblp-author    &8(5.40) &26.69 &2674721 &10(8.60) &27.25 &1089614 &10(9.50) &21.27 &696636 \\
dbpedia-genre    &5(2.85) &18.07 &546103 &4(2.85) &19.69 &121851 &4(3.05) &9.35 &147990 \\
dbpedia-team    &4(3.25) &16.25 &5699436 &4(3.30) &8.94 &1232615 &5(3.85) &24.11 &1260575 \\
discogs\_style    &9(3.30) &1.19 &10651 &7(3.80) &5.14 &13617 &25(7.20) &29.43 &19900 \\
edit-frwiktionary    &9(2.65) &0.18 &2476  &9(2.85) &5.52 &2204 &9(3.05) &1.28 &1946 \\
wiki-en-cat    &14(6.05) &29.28 &3640199 &13(7.50) &24.17 &553078  &14(8.75) &25.18 &706691 \\
	\hline
	\end{tabular}
	\end{scriptsize}
	\end{table}
	

The Constraint-Based Tabu Search procedure (Algorithm \ref{Algorithm_General}, line 6) is one key component of the proposed TSGR-MBBP algorithm. One of the main features of CBTS is that while unbalanced bicliques are allowed, the balance deviation of the explored bicliques must be no more than 2 (see Sections \ref{sec_prob_formul} and \ref{sec_tabu_search}) (this constraint is called unbalance constraint). To justify this specific unbalance constraint, we compare CBTS with two CBTS versions with different unbalance constraints. The first version (called ``CBTS$_{\Omega^\infty}$") removes the unbalance constraint and allows the procedure to visit any bicliques (lines 15-23 are removed from Algorithm \ref{algorithm_tabu_search}). The second version (named as ``CBTS$_{\Omega^1}$") introduces a more restrictive unbalance constraint -- the balance deviation is required to be no more than 1 after each iteration (i.e., change the repairing condition in line 15 to $|X|-|Y|>1$). We also used ``CBTS$_{\Omega^2}$" to denote the original CBTS procedure. As such, these three CBTS versions correspond to three restart algorithms searching  within the solution spaces $\Omega^2$, $\Omega^\infty$, and $\Omega^1$ respectively. Note that the version with absolute balanced constraint is not considered. In effect, if we repair the solution whenever $|X|-|Y|\neq 0$, the current solution can never be improved because the ``push" operator only imports one vertex to one vertex set in each iteration.

For this study, we used 13 instances selected from the two benchmark sets. We ran each CBTS version 20 trials to solve each instance under the same configuration mentioned in Section \ref{subsec_exp_protocoal}. Each trial was given a time limit of 60 seconds. The comparative results of this study are summarized in Table \ref{tbl_banalce_cmp}. We denote one restart of CBTS as one iteration here (one `while' loop, lines 10-20 in Algorithm \ref{Algorithm_General}). Column ``best(ave)" indicates the best and average balanced biclique size found by each algorithm over 20 runs. Column ``time" reports the average time to achieve the best balanced biclique size in all 20 runs. Column ``iter" reports the average number of restarts for 20 runs.

As for the solution quality, the original Constraint-Based Tabu Search (CBTS$_{\Omega^2}$) procedure dominates the other variants both in terms of best and average values. CBTS$_{\Omega^2}$ also performs the best concerning the average time of attaining the best solution for random graphs. As for the total number of iterations (column ``iter"), CBTS$_{\Omega^\infty}$ restarts more often than CBTS$_{\Omega^2}$ which on the other hand restarts more often than CBTS$_{\Omega^1}$. Obviously, a tighter unbalance constraint leads to more frequent calls to the repair procedure, thus less iterations under the same time limitation. Meanwhile, the results suggest that the strategy of incorporating unbalance constraint is a good trade-off between solution quality and number of iterations.

\subsection{Effectiveness of reduction methods}
	\begin{figure}	
	 \centering\scalebox{0.6}{\includegraphics[scale=1]{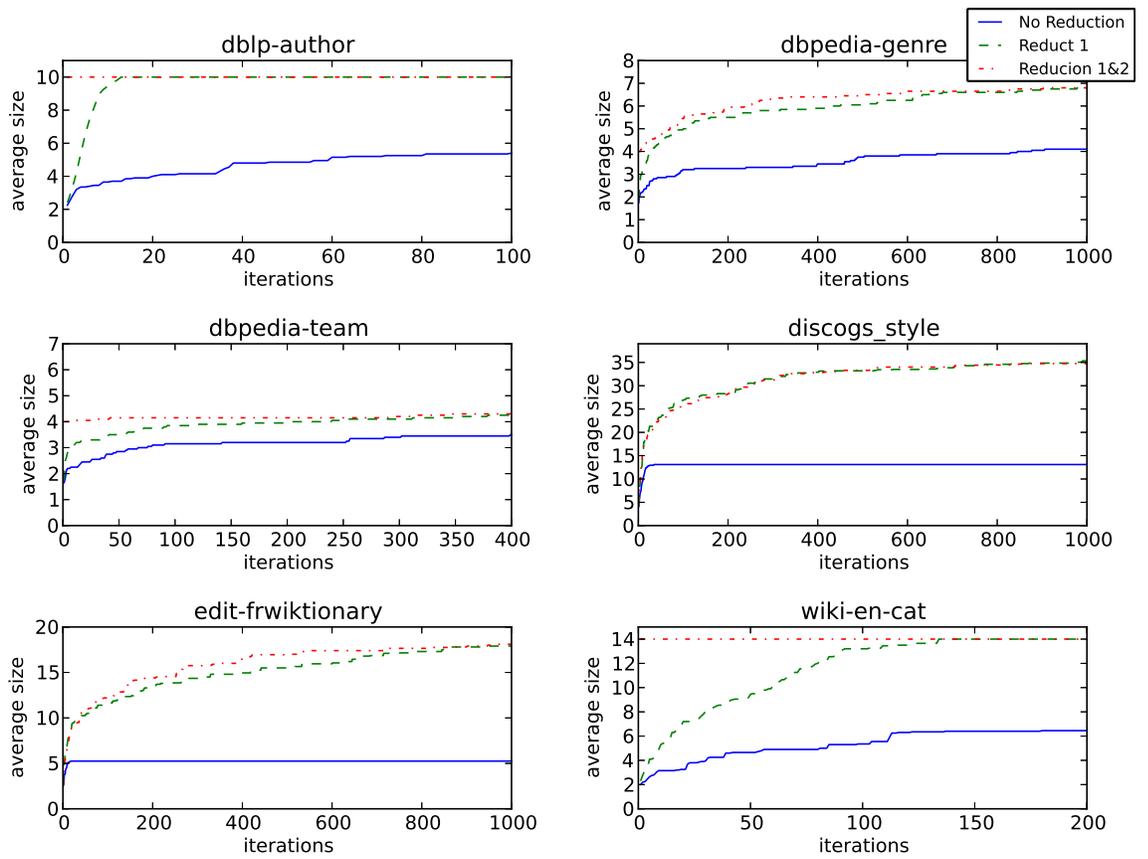}}
	 \centering\caption{The relations between the number of iterations and the average best sizes of 20 runs on 6 selected instances from KONECT. } \label{fig_iter-ave}
	\end{figure}

To gain a comprehensive understanding of the run-time behavior and efficiency of the two reduction methods, we show in this section an analysis of the convergence rate of three variants of the TSGR-MBBP algorithm:
\begin{itemize}
\item \textit{No Reduction:} The reduction procedure is disabled, i.e., lines 10-18 are removed from Algorithm \ref{Algorithm_General}.
\item \textit{Reduction 1:} Only the first reduction method (the $Peel$ method) is used. i.e., lines 12-18 are removed from Algorithm \ref{Algorithm_General}.
\item \textit{Reduction 1\&2:} Both reduction methods are used, i.e., the original TSGR-MBBP algorithm.
\end{itemize}

The variant with only the second reduction is not considered as the exact search will never be triggered without the $Peel$ procedure.

This study was based on 6 KONECT instances, ``dblp-author", ``dbpedia-genre", ``dbpedia-team", ``discog\_style", ``edit-frwikitionary" and ``wiki-en-cat" which are large enough with different levels of difficulty for TSGR-MBBP (the difficulty is estimated by the time consumption of TSGR-MBBP in Table \ref{tbl_large_ins}). We ran each algorithm variant 20 times to solve each instance with a time limit of 6 minutes per run. Again, we denote one restart of CBTS as one iteration. Figure \ref{fig_iter-ave} reports the relation between the number of iterations and the average best balanced size reached by each variant in 20 runs (abbreviated as `average size'). Considering the two variants with reduction can stop before reaching the time limit when the optimum is proven, we assume that the best size after termination is constantly the optimal size in this case. 

According to Figure \ref{fig_iter-ave}, in terms of the average result after the same number of iterations, the two variants using reduction always dominate the variant without reduction. Actually, ``No Reduction" converges so slowly that it even has difficulties in reaching half of the best-known size in the given time limit. Comparing ``Reduction 1" and ``Reduction 1\&2", for ``dblp-author", ``dbpedia-genre", ``dbpedia-team", ``edit\_fiwikitionaryand" and ``wiki-en-cat", ``Reduction 1\&2" always discovers solutions of high quality earlier. In particular, for two instances, ``dblp-author" and ``wiki-en-cat", ``Reduction 1\&2" reaches the optimal solution in the very first iteration. This is because for these graphs, the exact algorithm discovered the optimal solution in some of the connected subgraphs at the beginning of the search, which in turn enabled the \textit{peel} procedure to prune the graph to trivial size and thus proves the global optimality. Nevertheless, for ``discogs\_style", ``Reduction 1\&2" and ``Reduction 1" perform similarly. We also notice that the curves of ``Reduction 1" and ``Reduction 1\&2" meet sooner or later for all the instances. In a nutshell, the convergence rate is highly related to the instance under consideration, but in any case, both reduction methods accelerate the search procedure.

\section{Conclusions and perspectives}
\label{conclusions}

The Maximum	Balanced Biclique Problem is of great interest both theoretically and practically. We have presented an original tabu search combined with two dedicated graph reduction techniques for solving MBBP approximately. The proposed TSGR-MBBP algorithm is driven by a Constraint-Based Tabu Search (CBTS) procedure to retrieve high quality solutions from the current graph. CBTS employs the ``push" operator to explore relaxed search space including both balanced and unbalanced bicliques and imposes a specific unbalance constraint on explored solutions. Moreover, each time the lower bound is updated by CBTS, two reduction rules are used to prune the graph, which leads to a reduced search space for the following iterations. Specifically, the first reduction rule is based on removing unpromising vertices according to their degrees, while the second reduction rule removes small subgraphs using an exact search procedure. 

The TSGR-MBBP algorithm has been assessed on two benchmark sets: 30 random dense instances and 25 real-life large instances from the KONECT collection. For the random instances, TSGR-MBBP dominates existing state-of-the-art approaches EA/SM \citep{yuan2015new}, GL\_Greedy \citep{al2007defect} and CPLEX (version 12.6.1). Besides, new improved solutions (new lower bounds) were found by TSGR-MBBP for 10 out of the 30 instances. For the KONECT instances, TSGR-MBBP proved optimal solutions for 14 instances for the first time and found high quality solutions for the other instances. Experiments have also indicated that TSGR-MBBP performs better than CPLEX both in terms of solution quality and computational time. Besides, we have also noticed that the two reduction methods are able to prune a significant number of vertices for large sparse graphs. Additional experiments have demonstrated the effectiveness of the adopted unbalance constraint used by tabu search and confirmed that the combination of two reduction methods significantly accelerates the convergence of the search procedure.

This study can be extended in several directions. First, we only investigated two typical cases of the general ``push" operator which correspond to the ``add" and ``swap" moves. It would be interesting to study other customized moves based on the  ``push" operator. Second, as shown in our literature review, there are few exact algorithms for MBBP. It would be useful to design more elaborated exact algorithms. For this purpose, the exact algorithm introduced in this work (for the purpose of graph reduction) could be served as a base version for further improvement. It is also appealing to adapt more advanced exact clique algorithms like \citep{pattabiraman2013fast,san2016new} to MBBP. Finally, given the effectiveness of the reduction techniques on large-scale instances for MBBP, it would be interesting to investigate similar techniques in the context of other clique related problems like the maximum clique problem \cite{wu2015review} and the maximum $k$-plex problem \citep{balasundaram2011clique} for massive graphs.

\section*{Acknowledgment}
We'd like to thank the authors of \citep{yuan2015new} for sharing the code of their EA/SM algorithm. The work was partially supported by the PGMO (2014-0024H) project from the Jacques Hadamard Mathematical Foundation (Paris, France). Support for the first author of this work from the China Scholarship Council (2013-2017) is also acknowledged.


\appendix
\section{The exact search algorithm}
\label{apx_sec_exact_search}

The exact algorithm (Algorithm \ref{algorithm_exact_search} and Proc. \ref{proc_bbexpand}) used in TSGR-MBBP is adapted from the classical B\&B algorithm for the maximum clique problem \cite{carraghan1990exact}. Instead of starting from the trivial lower bound 0, our exact algorithm receives an initial lower bound $\omega$ (Algorithm \ref{algorithm_exact_search}, line 2), which is the best balanced size ever found in TSGR-MBBP. $(X^*,Y^*)$, the best biclique found by the exact algorithm, is initialized as a tuple of two empty sets (Algorithm \ref{algorithm_exact_search}, line 3). If there is no solution better than $\omega$ in the current subgraph, $(X^*, Y^*)$ remains empty even after the exact search. In such a case, the real optimal solution is discarded as we are only interested in solutions better than $\omega$. The exact algorithm calls a recursive procedure $bbexpand$ (Proc. \ref{proc_bbexpand}) to start the branch and bound search.

Unlike the original algorithm in \cite{carraghan1990exact}, which only builds one set that forms a clique, the $bbexpand$ procedure alternatively builds two sets $A$ and $B$ ($|A|=|B|$ or $|A| +1 = |B|$ ) such that $A$ and $B$ form a biclique. Sets $C_A$ and $C_B$ contain the candidate vertices that may be added to $A$ and $B$ respectively, i.e., ($C_A = \bigcap_{i\in B} N(i)$ and $C_B = \bigcap_{i\in A} N(i)$). Each invocation of $bbexpand$ recursively traversals the feasible bicliques containing $A$ and $B$ with all possible combinations of $C_A$ and $C_B$ examined. The procedure works as follow:

Firstly, if candidate set $C_A$ is empty, $bbexpand$ tries to update the current lower bound $lb$  (lines 1-5, Proc. \ref{proc_bbexpand}). As the cardinality of set $A$ is equal to or one less than that of $B$ in the input of $bbexpand$, $|A|$ is always the balanced size of biclique $(A,B)$ (or $(B,A)$). Then, if $C_A$ is not empty, $bbexpand$ enters a while loop (lines 6-14), where in each iteration, a vertex $i$ with minimum index is picked from $C_A$ (lines 9-10) to form a new solution with $A\cup \{i\}$ and $B$ while removing $i$ from $C_A$ at the same time. In the end of the iteration, $bbexpand$ is recursively called to enumerate all the feasible bicliques with the new sets of solution and new candidate sets ($C_A$ and $C_B\cap N(i)$). Note that the roles of $A$ and $B$ as well as $C_A$ and $C_B$ in the next level of recursive call (lines 11-14) are exchanged. This is because that, to meet the balance constraint, the biclique is built by alternatively introducing a vertex from $U$ and $V$. The $if$ part in lines 7-8 prunes the unnecessary expanding since, when $|A|+|C_A|$ is smaller than the current lower bound, there is no possibility to discover a better solution based on the given solution and candidate sets. This simple rule of pruning unnecessary enumeration is similar to the one used in \cite{carraghan1990exact}. In brief, every loop from line 6 to line 14 enumerates possible bicliques involving set $A$ with a newly selected vertex $i$ and set $B$.

\begin{algorithm}[H]
\begin{small}
\caption{Exact search algorithm}
\label{algorithm_exact_search}
\KwIn{Graph instance $G(U,V,E)$, initial lower bound $\omega$}
\KwOut{A biclique $(X^*, Y^*)$ with maximum balanced size}
\Begin{
$lb\gets \omega$ \tcc*[r]{Initialize the lower bound as $\omega$}
$(X^*,Y^*) \gets (\emptyset, \emptyset)$ \tcc*[r]{Initialize the best solution as empty sets}
bbexpand($G$, $\emptyset$, $\emptyset$, $U$, $V$)\;
} 
\Return{$(X^*, Y^*)$}

\end{small}
\end{algorithm}

\begin{procedure}
\begin{small}
	\caption{bbexpand($G$, $A$, $B$, $C_A$, $C_B$)}
	\label{proc_bbexpand}
	\KwIn{Graph instance $G=(U,V,E)$, $A$, $B$ - current sets that forms a biclique, $C_A$, $C_B$ - the sets of eligible vertices that can be added to $A$ and $B$ respectively.}
	\KwOut{The maximum balanced size $\omega$ in $G$, the biclique $(X^*,Y^*)$ with balanced size $\omega$.}
	\If{$|C_A| = 0$}{
		\If{$|A| > \omega$}{
			$lb \gets |A|$ \;
			Record current solution $(A,B)$ in $(X^*, Y^*)$\;
			\Return
		}
	}
	\While{$C_A \neq \emptyset$}{
		\If{$|A| + |C_A| \leq lb$}{
			\Return
		}
		$i \gets min\{i|i \in C_A\}$\;
		$ C_A \gets C_A\setminus\{i\}$\;
		\If{$|A| < |B|$}{
			bbexpand($G$,$A \cup \{i\}$, $B$, $C_A$, $C_B\cap N(i)$)	
		}
		\Else{ 					
			bbexpand($G$,$B$, $A \cup \{i\}$, $C_B\cap N(i)$, $C_A$)
		}
	}
	\Return
\end{small}
\end{procedure}

\end{document}